\documentclass[10pt, conference, compsocconf]{IEEEtran}
\usepackage{ifpdf}
\usepackage{cite}

\ifCLASSINFOpdf
   \usepackage[pdftex]{graphicx}
\else
   \usepackage[dvips]{graphicx}
\fi

\usepackage[cmex10]{amsmath}
\usepackage{algorithmic}
\usepackage{array}
\usepackage{mdwmath}
\usepackage{mdwtab}
\usepackage{eqparbox}

\usepackage[tight,footnotesize]{subfigure}
\usepackage[caption=false,font=footnotesize]{subfig}

\usepackage{url}

\begin{document}

\title{Recurrent Neural Networks Hardware Implementation on FPGA}


\author{\IEEEauthorblockN{Andre Xian Ming Chang, Berin Martini, Eugenio Culurciello}
\IEEEauthorblockA{Department of Electrical and Computer Engineering\\
Purdue University\\
West Lafayette, USA\\
\texttt{\{amingcha,berin,euge\}@purdue.edu}}
}



\maketitle

\begin{abstract}
Recurrent Neural Networks (RNNs) have the ability to retain memory and learn data sequences. Due to the recurrent nature of RNNs, it is sometimes hard to parallelize all its computations on conventional hardware. CPUs do not currently offer large parallelism, while GPUs offer limited parallelism due to sequential components of RNN models. In this paper we present a hardware implementation of Long-Short Term Memory (LSTM) recurrent network on the programmable logic Zynq 7020 FPGA from Xilinx. We implemented a RNN with $2$ layers and $128$ hidden units in hardware and it has been tested using a character level language model. The implementation is more than $21\times$ faster than the ARM Cortex-A9 CPU embedded on the Zynq 7020 FPGA. This work can potentially evolve to a RNN co-processor for future mobile devices.
\end{abstract}



\section{Introduction}

As humanity progresses into the digital era, more and more data is produced and distributed across the world. Deep Neural Networks (DNN) provides a method for computers to learn from this mass of data. This unlocks a new set of possibilities in computer vision, speech recognition, natural language processing and more. However, DNNs are computationally expensive, such that general processors consume large amounts of power to deliver desired performance. This limits the application of DNNs in the embedded world. Thus, a custom architecture optimized for DNNs provides superior performance per power and brings us a step closer to self-learning mobile devices.    

Recurrent Neural Networks (RNNs) are becoming an increasingly popular way to learn sequences of data 
\cite{sutskever2014sequence,cho2014learning,zaremba2014recurrent,graves2013speech}, and it has been shown to be successful in various applications, such as speech recognition \cite{graves2013speech}, machine translation \cite{sutskever2014sequence} and scene analysis \cite{byeon2015scene}. A combination of a Convolutional Neural Network (CNN) with a RNN can lead to fascinating results such as image caption generation \cite{vinyals2014show,mao2014explain,fang2014captions}. 

Due to the recurrent nature of RNNs, it is sometimes hard to parallelize all its computations on conventional hardware. General purposes CPUs do not currently offer large parallelism, while small RNN models do not get full benefit from GPUs. Thus, an optimized hardware architecture is necessary for executing RNNs models on embedded systems.

\begin{figure}[!t]
\centering
\includegraphics[width=3.2in]{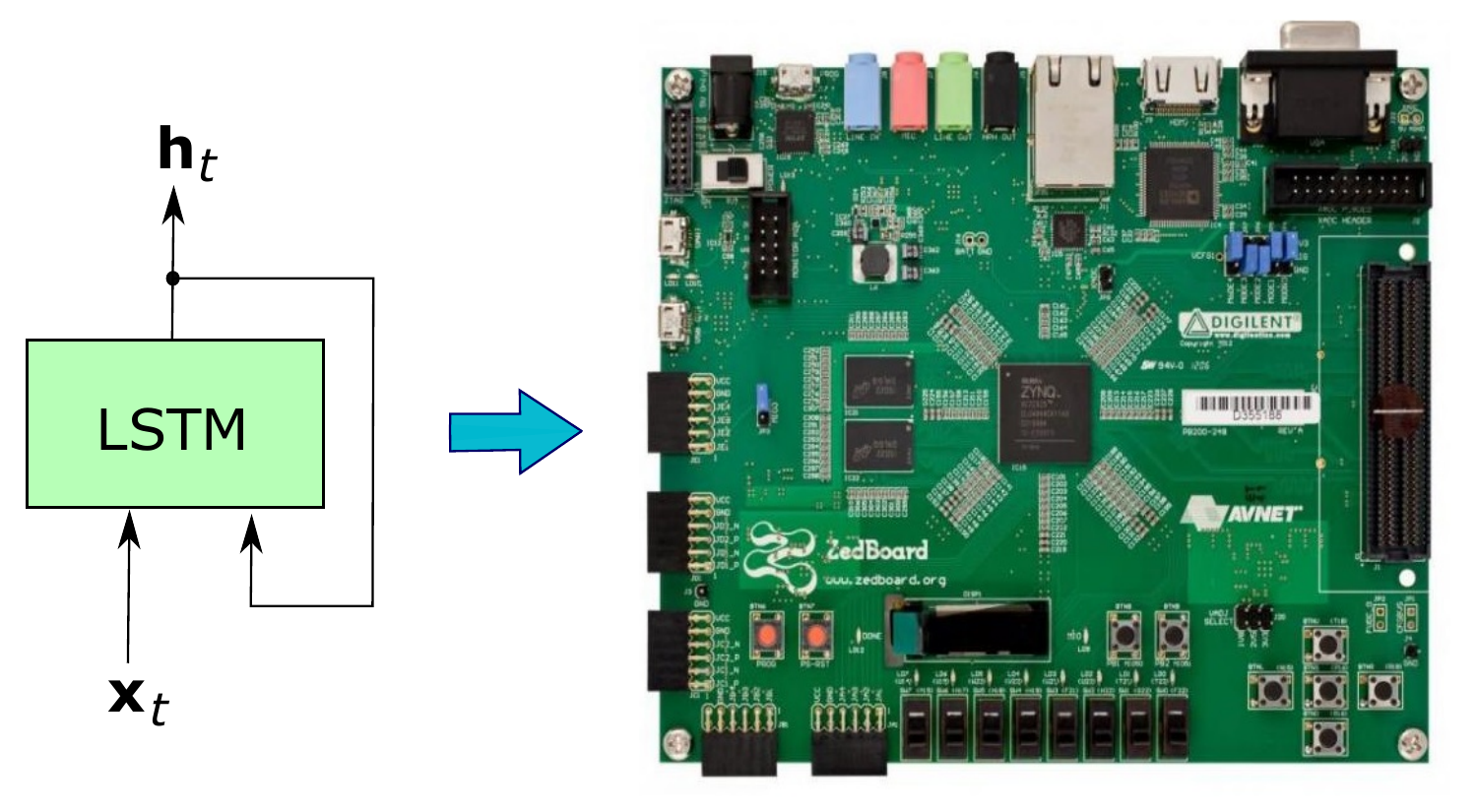}
\caption{The LSTM hardware was implemented using a Zedboard Zynq ZC7020.}
\label{fig:bigpic}
\end{figure}

Long Short Term Memory, or LSTM \cite{hochreiter1997long}, is a specific RNN architecture that implements a learned memory controller for avoiding vanishing or exploding gradients \cite{bengio1994learning}. The purpose of this paper is to present a LSTM hardware module implemented on the Zynq 7020 FPGA from Xilinx \cite{xilinx:zynq7000}. Figure \ref{fig:bigpic} shows an overview of the system. As proof of concept, the hardware was tested with a character level language model made with $2$ LSTM layers and $128$ hidden units. The next following sections present the background for LSTM, related work, implementation details of the hardware and driver software, the experimental setup and the obtained results.

\section{LSTM Background}
One main feature of RNNs are that they can learn from previous information. But the question is how far should a model remember, and what to remember. Standard RNN can retain and use recent past information \cite{schmidhuber2015deep}. But it fails to learn long-term dependencies. Vanilla RNNs are hard to train for long sequences due to vanishing or exploding gradients \cite{bengio1994learning}. This is where LSTM comes into play. LSTM is an RNN architecture that explicitly adds memory controllers to decide when to remember, forget and output. This makes the training procedure much more stable and allows the model to learn long-term dependencies \cite{hochreiter1997long}. 

There are some variations on the LSTM architecture. One variant is the LSTM with peephole introduced by \cite{gers2000recurrent}. In this variation, the cell memory influences the input, forget and output gates. Conceptually, the model peeps into the memory cell before deciding whether to memorize or forget. In \cite{cho2014learning}, input and forget gate are merged together into one gate. There are many other variations such as the ones presented in  \cite{sak2014long} and \cite{otte2014dynamic}. All those variations have similar performance as shown in \cite{greff2015lstm}. 

The LSTM hardware module that was implemented focuses on the LSTM version that does not have peepholes, which is shown in figure \ref{fig:lstm}. This is the vanilla LSTM \cite{graves2005framewise}, which is characterized by the following equations:

\begin{figure}[!t]
\centering
\includegraphics[width=3.4in]{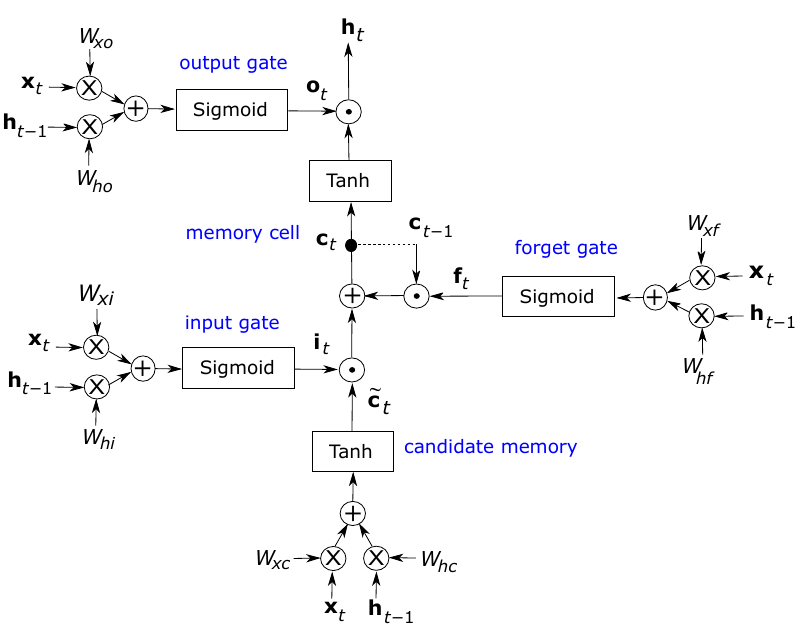}
\caption{The vanilla LSTM architecture that was implemented in hardware. $\otimes$ represents matrix-vector multiplication and $\odot$ is element-wise multiplication.}
\label{fig:lstm}
\end{figure}

\begin{equation} \label{eq:1}
\mathbf{i}_{t} = \sigma (W_{xi}\mathbf{x}_{t} + W_{hi}\mathbf{h}_{t-1} + \mathbf{b}_{i})
\end{equation}
\begin{equation} \label{eq:2}
\mathbf{f}_{t} = \sigma (W_{xf}\mathbf{x}_{t} + W_{hf}\mathbf{h}_{t-1} + \mathbf{b}_{f})
\end{equation}
\begin{equation} \label{eq:3}
\mathbf{o}_{t} = \sigma (W_{xo}\mathbf{x}_{t} + W_{ho}\mathbf{h}_{t-1} + \mathbf{b}_{o})
\end{equation}
\begin{equation} \label{eq:4}
\mathbf{\tilde{c}}_{t} = \mathrm{tanh}(W_{xc}\mathbf{x}_{t} + W_{hc}\mathbf{h}_{t-1} + \mathbf{b}_{c})
\end{equation}
\begin{equation} \label{eq:5}
\mathbf{c}_{t} = \mathbf{f}_{t} \odot \mathbf{c}_{t-1} + \mathbf{i}_{t} \odot \mathbf{\tilde{c}}_{t}
\end{equation}
\begin{equation} \label{eq:6}
\mathbf{h}_{t} = \mathbf{o}_{t} \odot \mathrm{tanh}(\mathbf{c}_{t})
\end{equation}

where $\sigma$ is the logistic sigmoid function, $\odot$ is element wise multiplication, $\mathbf{x}$ is the input vector of the layer, $W$ is the model parameters, $\mathbf{c}$ is memory cell activation, $\mathbf{\tilde{c}}_{t}$ is the candidate memory cell gate, $\mathbf{h}$ is the layer output vector. The subscript $t-1$ means results from the previous time step. The $\mathbf{i}$, $\mathbf{f}$ and $\mathbf{o}$ are respectively input, forget and output gate. Conceptually, these gates decide when to remember or forget an input sequence, and when to respond with an output. The combination of two matrix-vector multiplications and a non-linear function, $f(W_{x}\mathbf{x}_{t} + W_{h}\mathbf{h}_{t-1} + \mathbf{b})$, extracts information from the \emph{input} and \emph{previous output} vectors. This operation is referred as gate. 

One needs to train the model to get the parameters that will give the desired output. In simple terms, training is an iterating process in which training data is fed in and the output is compared with a target. Then the model needs to backpropagate the error derivatives to update new parameters that minimize the error. This cycle repeats until the error is small enough \cite{bishop2006pattern}. Models can become fairly complex as more layers and more different functions are added. For the LSTM case, each module has four gates and some element-wise operations. A deep LSTM network would have multiple LSTM modules cascaded in a way that the output of one layer is the input of the following layer.

\section{Related Work}
Co-processors for accelerating computer vision algorithms and CNNs have been implemented on FPGAs. A system that can perform recognition on mega-pixel images in real-time is presented in \cite{farabet2010hardware}. A similar architecture for general purpose vision algorithms called \emph{neuFlow} is described in \cite{farabet2011neuflow}. \emph{neuFlow} is a scalable architecture composed by a grid of operation modules connected with an optimized data streaming network. This system can achieve speedups up to $100\times$ in end-to-end applications.

An accelerator called \emph{nn-X} for deep neural networks is described in \cite{dundar2013accelerating, jin2014efficient, dundar2014memory,  gokhale2014240}. \emph{nn-X} is a high performance co-processor implemented on FPGA. The design is based on computational elements called collections that are capable of performing convolution, non-linear functions and pooling. The accelerator efficiently pipelines the collections achieving up to $240$\,G-op/s.

RNNs are different from CNNs in the context that they require a different arrangement of computation modules. This allows different hardware optimization strategies that should be exploited. A LSTM learning algorithm using Simultaneous Perturbation Stochastic Approximation (SPSA) for hardware friendly implementation was described in \cite{tavcar2013transforming}. The paper focuses on transformation of the learning phase of LSTM for FPGA.
 
Another FPGA implementation that focus on standard RNN is described by \cite{lifpga}. Their approach was to unfold the RNN model into a fixed number of timesteps $B$ and compute them in parallel. The hardware architecture is composed of a hidden layer module and duplicated output layer modules. First, the hidden layer serially processes the input $\mathbf{x}$ for $B$ timesteps. Then, with the results of the hidden layer, the duplicated logic computes output $\mathbf{h}$ for $B$ timesteps in parallel.

This work presents a different approach of implementing RNN in FPGA, focusing the LSTM architecture. It is different than \cite{lifpga}, in the sense that it uses a single module that consumes input $\mathbf{x}$ and previous output $\mathbf{h}_{t-1}$ simultaneously.

\section{Implementation}
\subsection{Hardware}
The main operations to be implemented in hardware are matrix-vector multiplications and non-linear functions (hyperbolic tangent and logistic sigmoid). Both are modifications of the modules presented in \cite{gokhale2014240}. For this design, the number format of choice is Q8.8 fixed point. The matrix-vector multiplication is computed by a Multiply ACcumulate (MAC) unit, which takes two streams: vector stream and weight matrix row stream. The same vector stream is multiplied and accumulated with each weight matrix row to produce an output vector with same size of the weight's height. The MAC is reset after computing each output element to avoid accumulating previous matrix rows computations. The bias $\mathbf{b}$ can be added in the multiply accumulate by adding the bias vector to the last column of the weight matrix and adding an extra vector element set to unity. This way there is no need to add extra input ports for the bias nor add extra pre-configuration step to the MAC unit. The results from the MAC units are added together. The adder's output goes to an element wise non-linear function, which is implemented with linear mapping. 

\begin{figure}[!t]
\centering
\includegraphics[width=2.9in]{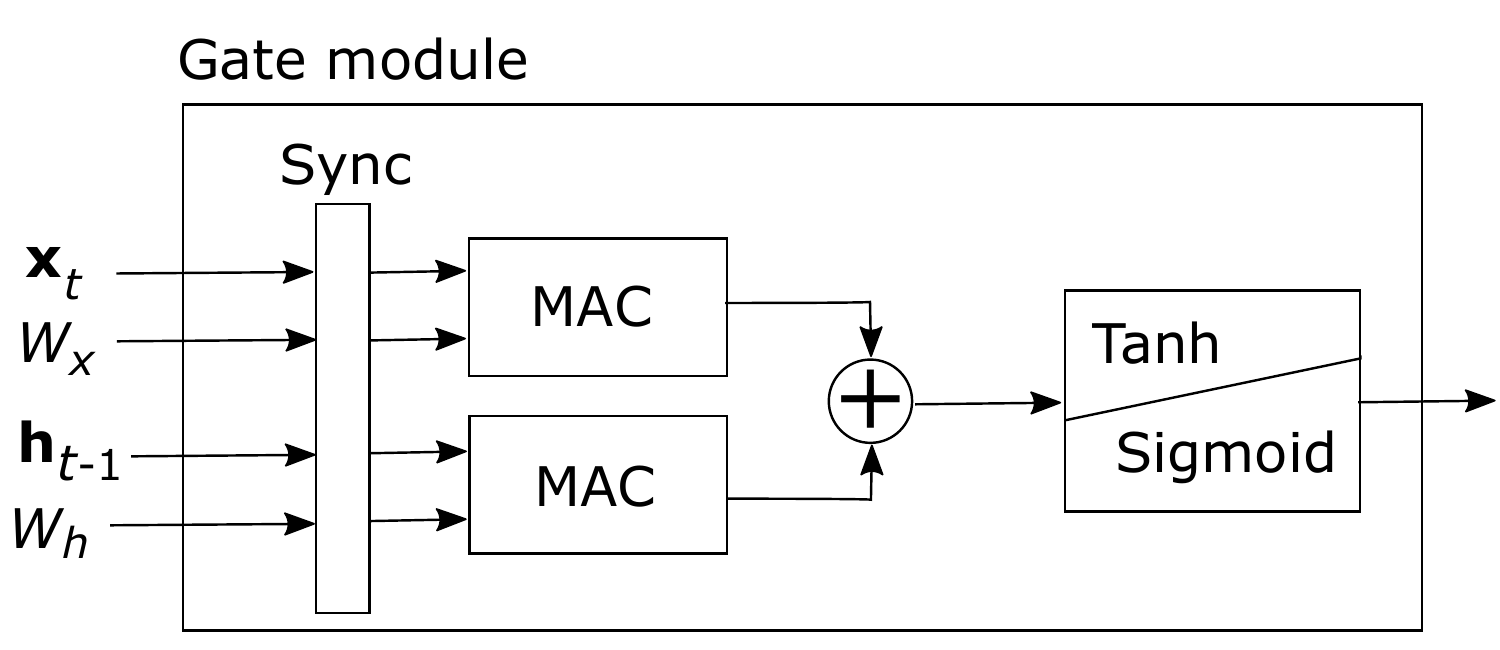}
\caption{The main hardware module that implements the LSTM gates. The non-linear module can be configured to be a tanh or logistic sigmoid.}
\label{fig:neuron}
\end{figure}

The non-linear function is segmented into lines $y=ax+b$, with $x$ limited to a particular range. The values of $a$, $b$ and $x$ range are stored in configuration registers during the configuration stage. Each line segment is implemented with a MAC unit and a comparator. The MAC multiplies $a$ and $x$ and accumulates with $b$. The comparison between the input value with the line range decides whether to process the input or pass it to the next line segment module. The non-linear functions were segmented into 13 lines, thus the non-linear module contains 13 pipelined line segment modules. 
The main building block of the implemented design is the gate module as shown in figure \ref{fig:neuron}.

The implemented module uses Direct Memory Access (DMA) ports to stream data in and out. The DMA ports use valid and ready handshake. Because the DMA ports are independent, the input streams are not synchronized even when the module activates the ports at same the time. Therefore, a stream synchronizing module is needed. The sync block is a buffer that caches some streaming data until all ports are streaming. When the last port starts streaming, the sync block starts to output synchronized streams. This ensures that vector and matrix row elements that goes to MAC units are aligned. 
 
The gate module in figure \ref{fig:neuron} also contains a rescale block that converts $32$ bit values to $16$ bit values. The MAC units perform $16$ bit multiplication that results into $32$ bit values. The addition is performed using $32$ bit values to preserve accuracy.

\begin{figure}[!t]
\centering
\includegraphics[width=3.0in]{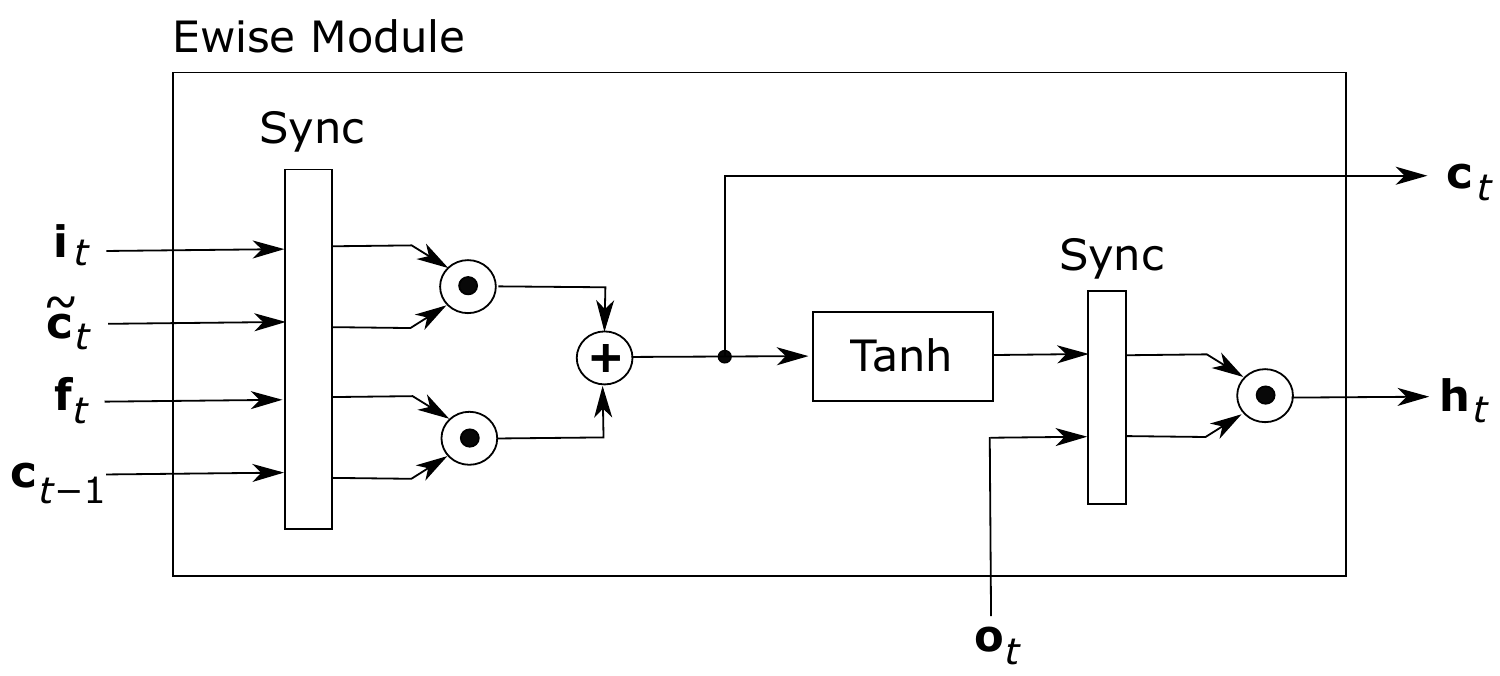}
\caption{The module that computes the $\mathbf{c}_{t}$ and $\mathbf{h}_{t}$ from the results of the gates. $\odot$ is element-wise multiplication. }
\label{fig:laststage}
\end{figure}

All that is left are some element wise operations to calculate $\mathbf{c}_{t}$ and $\mathbf{h}_{t}$ in equations \ref{eq:5} and \ref{eq:6}. To do this, extra multipliers and adders were added into a separate module shown in figure \ref{fig:laststage}. 

\begin{figure}[!t]
\centering
\includegraphics[width=2.7in]{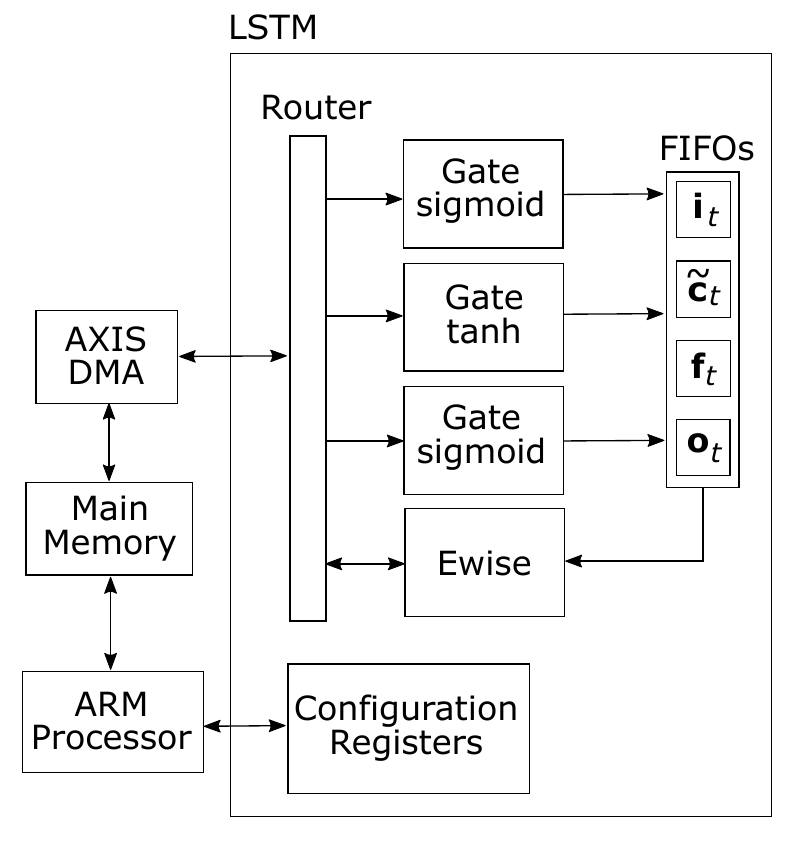}
\caption{The LSTM module block diagram. It is mainly composed of three gates and one ewise stage module. }
\label{fig:lstm_module}
\end{figure}

The LSTM module uses three blocks from figure \ref{fig:neuron} and one from figure \ref{fig:laststage}. The gates are pre-configured to have a non-linear function (tanh or sigmoid). The LSTM module is shown in figure\,\,\ref{fig:lstm_module}.

The internal blocks are controlled by a state machine to perform a sequence of operations. The implemented design uses four $32$ bit DMA ports. Since the operations are done in $16$ bit, each DMA port can transmit two $16$ bit streams. The weights $W_{x}$ and $W_{h}$ are concatenated in the main memory to exploit this feature. The streams are then routed to different modules depending on the operation to be performed. With this setup, the LSTM computation was separated into three sequential stages: 

\begin{enumerate}
\item Compute $\mathbf{i}_{t}$ and $\mathbf{\tilde{c}}_{t}$.
\item Compute $\mathbf{f}_{t}$ and $\mathbf{o}_{t}$.
\item Compute $\mathbf{c}_{t}$ and $\mathbf{h}_{t}$.
\end{enumerate}
 
In the first and second stage, two gate modules ($4$ MAC units) are running in parallel to generate two internal vectors ($\mathbf{i}_{t}$, $\mathbf{\tilde{c}}_{t}$, $\mathbf{f}_{t}$ and $\mathbf{o}_{t}$), which are stored into a First In First Out (FIFO) for the next stages. The ewise module consumes the FIFO vectors to output the $\mathbf{h}_{t}$ and $\mathbf{c}_{t}$ back to main memory. After that, the module waits for new weights and new vectors, which can be for the next layer or next time step. The hardware also implements an extra matrix-vector multiplication to generate the final output. This is only used when the last LSTM layer has finished its computation.

This architecture was implemented on the Zedboard \cite{Avnet:zedboard}, which contains the Zynq-7000 SOC XC7Z020. The chip contains Dual ARM Cortex-A9 MPCore, which is used for running the LSTM driver C code and timing comparisons. The hardware utilization is shown in table \ref{tab:hw}. The module runs at $142$\,MHz and the total on-chip power is $1.942$\,W.

\begin{table}[ht]
\caption{FPGA hardware resource utilization for Zynq ZC7020.}
\label{tab:hw}
\begin{center}
\begin{tabular}{l r r}
\\
\multicolumn{1}{c}{\bf Components } &\multicolumn{1}{c}{\bf Utilization [ / ] }  &\multicolumn{1}{c}{\bf Utilization [\%] }
\\ \hline \\
FF         & 12960       & 12.18            \\
LUT        & 7201        & 13.54            \\
Memory LUT & 426         & 2.45             \\
BRAM       & 16          & 11.43            \\
DSP48      & 50          & 22.73            \\
BUFG       & 1           & 3.12            
\end{tabular}
\end{center}
\end{table}

\subsection{Driving Software}
The control and testing software was implemented with C code. The software populates the main memory with weight values and input vectors, and it controls the hardware module with a set of configuration registers.

The weight matrix have an extra element containing the bias value in the end of each row. The input vector contains an extra unity value so that the matrix-vector multiplication will only add the last element of the matrix row (bias addition). Usually the input vector $\mathbf{x}$ size can be different from the output vector $\mathbf{h}$ size. Zero padding was used to match both the matrix row size and vector size, which makes stream synchronization easier.

Due to the recurrent nature of LSTM, $\mathbf{c}_{t}$ and $\mathbf{h}_{t}$ becomes the $\mathbf{c}_{t-1}$ and $\mathbf{h}_{t-1}$ for the next time step. Therefore, the input memory location for $\mathbf{c}_{t-1}$ and $\mathbf{h}_{t-1}$ is the same for the output $\mathbf{c}_{t}$ and $\mathbf{h}_{t}$. Each time step $\mathbf{c}$ and $\mathbf{h}$ are overwritten. This is done to minimize the number of memory copies done by the CPU. To implement a multi-layer LSTM, the output of the previous layer $\mathbf{h}_{t}$ was copied to the $\mathbf{x}_{t}$ location of the next layer, so that $\mathbf{h}_{t}$ is preserved in between layers for error measurements. This feature was removed for profiling time. The control software also needs to change the weights for different layers by setting different memory locations in the control registers.

\section{Experiments}
The training script by Andrej Karpathy of the character level language model was written in Torch7. The code can be downloaded from Github\footnote{\url{https://github.com/karpathy/char-rnn}}. Additional functions were written to transfer the trained parameters from the Torch7 code to the control software.

The Torch7 code implements a character level language model, which predicts the next character given a previous character. Character by character, the model generates a text that looks like the training data set, which can be a book or a large internet corpora with more than $2$\,MB of words. For this experiment, the model was trained on a subset of Shakespeare's work. The batch size was $50$, the training sequence was $50$ and learning rate was $0.002$. The model is expected to output Shakespeare look like text. 

The Torch7 code implements a $2$ layer LSTM with hidden layer size $128$ (weight matrix height). The character input and output is a $65$ sized vector one-hot encoded. The character that the vector represents is the index of the only unity element. The predicted character from last layer is fed back to input $\mathbf{x}_{t}$ of first layer for following time step.

For profiling time, the Torch7 code was ran on other embedded platforms to compare the execution time between them. One platform is the Tegra K1 development board, which contains quad-core ARM Cortex-A15 CPU and Kepler GPU 192 Cores. The Tegra's CPU was clocked at maximum frequency of $2320.5$\,MHz. The GPU was clocked at maximum of $852$\,MHz. The GPU memory was running at $102$\,MHz.

Another platform used is the Odroid XU4, which has the Exynos5422 with four high performance Cortex-A15 cores and four low power Cortex-A7 cores (ARM big.LITTLE technology). The low power Cortex-A7 cores was clocked at $1400$\,MHz and the high performance Cortex-A15 cores was running at $2000$\,MHz.

The C code LSTM implementation was ran on Zedboard's dual ARM Cortex-A9 processor clocked at $667$\,MHz. Finally, the hardware was ran on Zedboard's FPGA clocked at $142$\,MHz. 

\section{Results}
\subsection{Accuracy}
The number of weights of some models can be very large. Even our small model used almost $530$\,KB of weights. Thus it makes sense to compress those weights into different number formats for a throughput versus accuracy trade off. The use of fixed point Q8.8 data format certainly introduces rounding errors. Then one may raise the question of how much these errors can propagate to the final output. Comparing the results from the Torch7 code with the LSTM module's output for same $\mathbf{x}_{t}$ sequence, the average percentage error for the $\mathbf{c}_{t}$ was $3.9\%$ and for $\mathbf{h}_{t}$ was $2.8\%$. Those values are average error of all time steps. The best was $1.3\%$ and the worse was $7.1\%$. The recurrent nature of LSTM did not accumulate the errors and on average it stabilized at a low percentage.

\begin{figure}[!t]
\centering
\includegraphics[width=3.4in,height=2.8in]{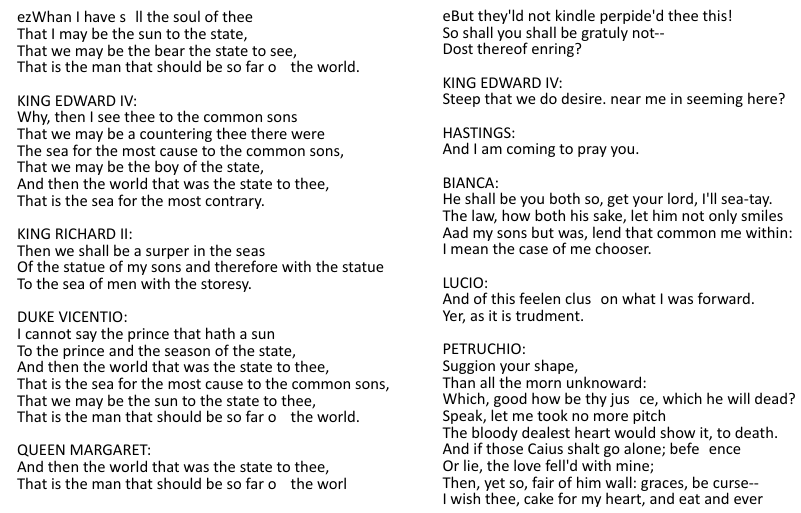}
\caption{On the left side is the output text from the LSTM hardware. On the right side is text from CPU implementation. The model predict each next character based on the previous characters. The model only gets the first character (seed) and generates the entire text character by character.}
\label{fig:text}
\end{figure}

The text generated by sampling $1000$ characters (timestep $t=1$ to $1000$) is shown in figure\,\ref{fig:text}. On the left is text output from FPGA and on the right is text from the CPU implementation. The result shows that the LSTM model was able to generate personage dialog, just like in one of Shakespeare's book. Both implementations displayed different texts, but same behavior. 

\subsection{Memory Bandwidth}
The Zedboard Zynq ZC7020 platform has $4$ Advanced eXtensible Interface (AXI) DMA ports available. Each is ran at $142$\,MHz and send packages of $32$\,bits. This allows aggregate bandwidth up to $3.8$\,GB/s full-duplex transfer between FPGA and external DDR3. 

At $142$\,MHz, one LSTM module is capable of computing $388.8$\,M-ops/s and uses simultaneously $4$ AXI DMA ports for streaming weight and vector values. During the peak memory usage, the module requires $1.236$\,GB/s of memory bandwidth. The high memory bandwidth requirements poses a limit to the number of LSTM modules that can be ran on parallel. To replicated LSTM module, it is required higher memory bandwidth or to introduce internal memory to lower requirements of external DDR3 memory usage. 

\subsection{Performance}
\begin{figure}[!t]
\centering
\includegraphics[width=3.5in,height=1.8in]{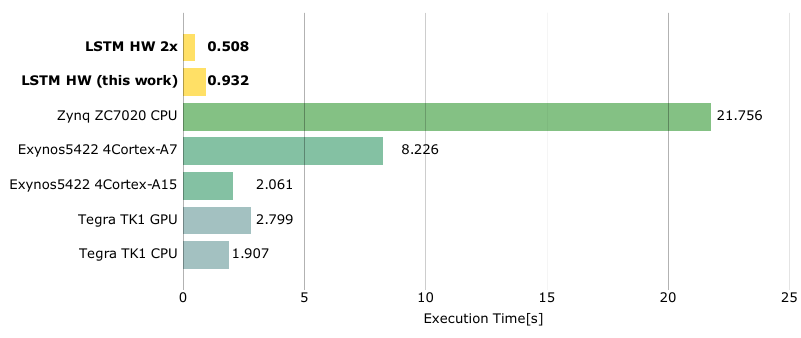}
\caption{Execution time of feedforward LSTM character level language model on different embedded platforms (the lower the better).}
\label{fig:exectime}
\end{figure}

Figure \ref{fig:exectime} shows the timing results. One can observe that the implemented hardware LSTM was significantly faster than other platforms, even running at lower clock frequency of $142$\,MHz (Zynq ZC7020 CPU uses $667$\,MHz). Scaling the implemented design by replicating the number of LSTM modules running in parallel will provide faster speed up. Using 2 LSTM cells in parallel can be $16\times$ faster than Exynos5422 on quad-core ARM Cortex-A7. 

\begin{figure}[!t]
\centering
\includegraphics[width=3.4in,height=1.8in]{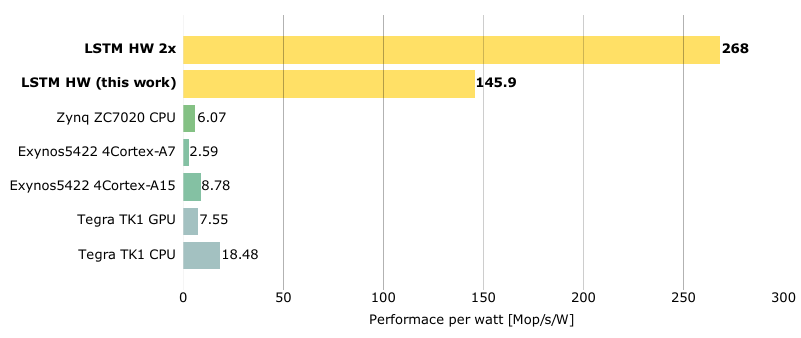}
\caption{Performance per unit power of different embedded platforms (the higher the better).}
\label{fig:perfW}
\end{figure}

In one LSTM layer of size $128$, there are $132.1$ K-ops. Multiplying this by number of samples and number of layers for the experimental application ($2000$) gives the total number of operations $264.4$ M-ops. The execution time divided by number of operations gives performance (ops/s). The power consumption of each platform was measured. Figure \ref{fig:perfW} shows the performance per unit power.  


The GPU performance was slower because of the following reasons. The model is too small for getting benefit from GPU, since the software needs to do memory copies. This is confirmed by running the same Torch7 code on a MacBook PRO 2016. The CPU of the MacBook PRO 2016 executed the Torch7 code for character level language model in $0.304$\,s, whereas the MacBook PRO 2016's GPU executed the same test in $0.569$\,s.

\section{Conclusion}
Recurrent Neural Networks have recently gained popularity due to the success from the use of Long Short Term Memory architecture in many applications, such as speech recognition, machine translation, scene analysis and image caption generation.

This work presented a hardware implementation of LSTM module. The hardware successfully produced Shakespeare-like text using a character level model. Furthermore, the implemented hardware showed to be significantly faster than other mobile platforms. This work can potentially evolve to a RNN co-processor for future devices, although further work needs to be done. The main future work is to optimize the design to allow parallel computation of the gates. This involves designing a parallel MAC unit configuration to perform the matrix-vector multiplication.

\subsubsection*{Acknowledgments}
This work is supported by Office of Naval Research (ONR) grants 14PR02106-01 P00004 and MURI N000141010278 and National Council for the Improvement of Higher Education (CAPES) through Brazil scientific Mobility Program (BSMP).
We would like to thank Vinayak Gokhale for the discussion on implementation and hardware architecture and also thank Alfredo Canziani, Aysegul Dundar and Jonghoon Jin for the support.
We gratefully appreciate the support of NVIDIA Corporation with the donation of GPUs used for this research.

\bibliographystyle{IEEEtran}
\bibliography{IEEEabrv,IEEEexample}

\end{document}